\documentclass[lettersize,journal]{IEEEtran}
\usepackage{amsmath,amsfonts}
\usepackage{algorithmic}
\usepackage{algorithm}
\usepackage{array}
\usepackage[caption=false,font=normalsize,labelfont=sf,textfont=sf]{subfig}
\usepackage{textcomp}
\usepackage{stfloats}
\usepackage{url}
\usepackage{verbatim}
\usepackage{graphicx}
\usepackage{cite}

\usepackage{booktabs}   
\usepackage{caption}    

\usepackage{cite}                       
\usepackage[hidelinks,colorlinks,
            linkcolor=black,
            citecolor=blue,
            urlcolor=blue]{hyperref}     
\usepackage[nameinlink,noabbrev]{cleveref}

\usepackage{xcolor}    

\begin{document}

\title{Spectral entropy prior-guided deep feature fusion architecture for magnetic core loss}

\author{
Cong Yao,Chunye Gong and Jin Zhang 

\thanks{The corresponding author 
is Jin Zhang .Jin Zhang and Cong Yao are  from the Changsha University of Science and Technology,Changsha 410114, China (e-mail: jzhang@csust.edu.cn,ycking@stu.csust.edu.cn).Chunye Gong is from  the National University of Defense Technology,Changsha 410073, China(e-mail: gongchunye@nudt.edu.cn);
}
}

\markboth{
IEEE Transactions on Power Electronics,~Vol.~14, No.~8, August~2025
}%
{Shell \MakeLowercase{\textit{et al.}}: A Sample Article Using IEEEtran.cls for IEEE Journals}


\maketitle

\begin{abstract}
Accurate core loss modeling is critical for the design of high-efficiency power electronic systems. Traditional core loss modeling methods have obvious limitations in prediction accuracy. To advance this field, the IEEE Power Electronics Society launched the MagNet Challenge in 2023, the first international competition focused on power electronics design methods, aiming to uncover the complex loss patterns of power magnetic components through a data-driven paradigm. Although pure data-driven models demonstrate significant advantages in fitting performance, their interpretability and cross-distribution generalization capabilities remain insufficient. To address this, this paper proposes a hybrid model, SEPI-TFPNet, that integrates empirical models and deep learning. The model consists of two parts: the physical prior submodule employs a spectral entropy discrimination mechanism to switch to the most suitable empirical model under different excitation waveforms, enabling explicit modeling of operational diversity; The data-driven submodule integrates convolutional neural networks, multi-head attention mechanisms, and bidirectional long short-term memory networks to deeply extract flux density time-series features. It also introduces an adaptive feature fusion module to effectively address feature interaction and information integration issues in multimodal data. Based on the MagNet dataset of various magnetic materials, this paper systematically evaluates the proposed method and compares it with 21 representative models included in the 2023 challenge and three advanced methods proposed in 2024–2025. The results show that the proposed method outperforms the state-of-the-art models, verifying the model's excellent modeling accuracy and robustness. 
\end{abstract}


\section{Introduction}
\IEEEPARstart{M}{agnetic}   material loss has long been a key bottleneck limiting the performance of high-frequency applications. The core loss process is influenced by various nonlinear factors, including the non-sinusoidal characteristics of the excitation waveform and environmental temperature drift. Therefore, developing an accurate power magnetic loss modeling method is a key focus and challenge in current research.

Currently, core loss modeling mainly relies on curve fitting methods and neural network methods. Curve fitting methods include the Steinmetz equation (SE) \cite{1,10} and its improved methods, but these methods have obvious limitations in practical applications.

In recent years, data-driven modeling methods have made significant progress, particularly in the field of neural networks (NNs). The empirical model-enhanced neural network (EMPINN) \cite{29} was proposed, which introduces a novel empirical model and integrates it into the FNN model, significantly improving the accuracy of loss prediction. However, this method cannot effectively handle multimodal data because it only uses scalar features as input, making it difficult to extract the spatio-temporal correlation information embedded in time series data. Sequence information typically contains temporal features and spatio-temporal relationships, while scalar data may contain key global features . Therefore, comprehensive multimodal processing of both is crucial. PI-MFF-CN \cite{28} was proposed, using CNN to process sequence data and FCNN to process scalar data, effectively addressing the issue of multi-source information processing. However, it achieves feature fusion through simple concatenation, failing to fully leverage the complementarity between different modalities. Effective fusion of multi-modal features is critical and directly impacts core loss prediction. Knowledge-based artificial neural networks (KANN) \cite{10} were developed, using the output of the SE function or iGSE function as one of the inputs to the feedforward neural network (FNN) \cite{7}, achieving high accuracy even with limited training data. However, its empirical model embedding is relatively coarse, failing to adapt the empirical model to different waveforms, leading to prediction errors under complex excitation conditions.

Given the limitations of previous research and methods, this paper proposes a hybrid magnetic core loss prediction model based on an improved CNN , Bi-LSTM, attention mechanism , and empirical model, namely SEPI-TFPNet. Its main contributions are summarized as follows:

1) This paper proposes a high-accuracy hybrid SEPI-TFPNet model that employs a precisely designed feature extraction and representation mechanism. Extensive experiments demonstrate that the proposed model achieves significantly higher prediction accuracy than existing state-of-the-art methods under large-scale scenarios, confirming its superior performance.

2) A physics-informed prediction framework constrained by spectral entropy is constructed. Compared with purely data-driven models, the proposed approach substantially improves physical consistency and model interpretability, while effectively suppressing prediction fluctuations and enhancing robustness.

3) This work is the first to introduce an adaptive multimodal feature fusion mechanism for core loss prediction, enabling dynamic coordination and complementary representation between sequence features and scalar physical features, thereby resolving the long-standing issue of multimodal feature inconsistency.

4) The proposed SEPI-TFPNet framework integrates deep representation learning, physics-guided priors, and explicit feature extraction. This collaborative design not only ensures high prediction accuracy on large datasets but also effectively suppresses extreme-sample errors, significantly improving model generalization. Extensive comparative experiments further validate the enhanced generalization capability of the proposed model.

The structure of this paper is as follows: Section 1 provides an overview of the introduction, Section 2 outlines the methods and model architecture of this paper, Section 3 discusses the experimental results, and Section 4 presents the conclusions of this paper.

\begin{figure*}[t]
    \centering
    \includegraphics[width=\linewidth]{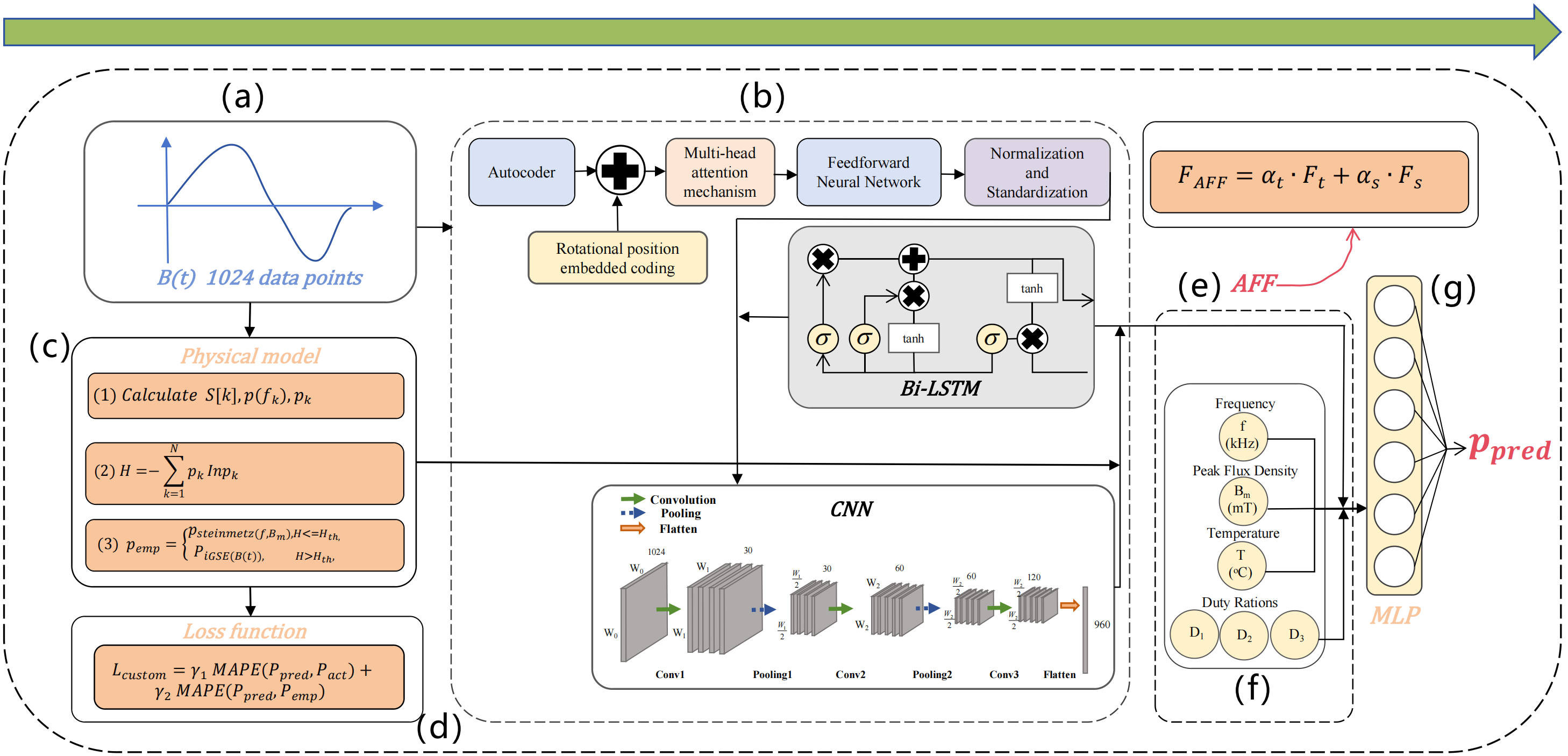}
    \caption{SEPI-TFPNet model architecture consists of:(a) an input layer that receives 1,024-point time-series data;(b) a deep learning subnetwork that processes signals through hierarchical feature extraction and representation;(c) a physics-informed empirical prior model constructed based on spectral-entropy discrimination;(d) a customized loss function that jointly constrains the empirical model and the data-driven model;(e) an adaptive multimodal fusion module that integrates time-series features and scalar physical features;(f) an explicit physics-prior feature channel that complements the learned deep representations;(g) a three-layer MLP-based core-loss prediction output layer.}
    \label{fig:1}
\end{figure*}

\begin{figure}[!t]
    \centering
    \includegraphics[width=0.65\linewidth]{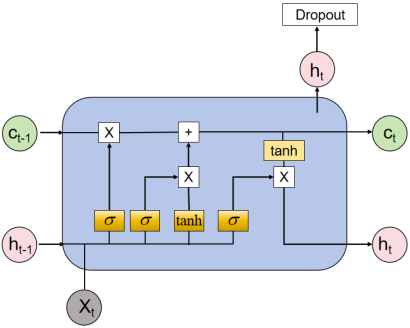}
    \caption{Traditional single LSTM structure}
    \label{fig:2}
\end{figure}

\begin{figure}[!t]
    \centering
    \includegraphics[width=0.75\linewidth]{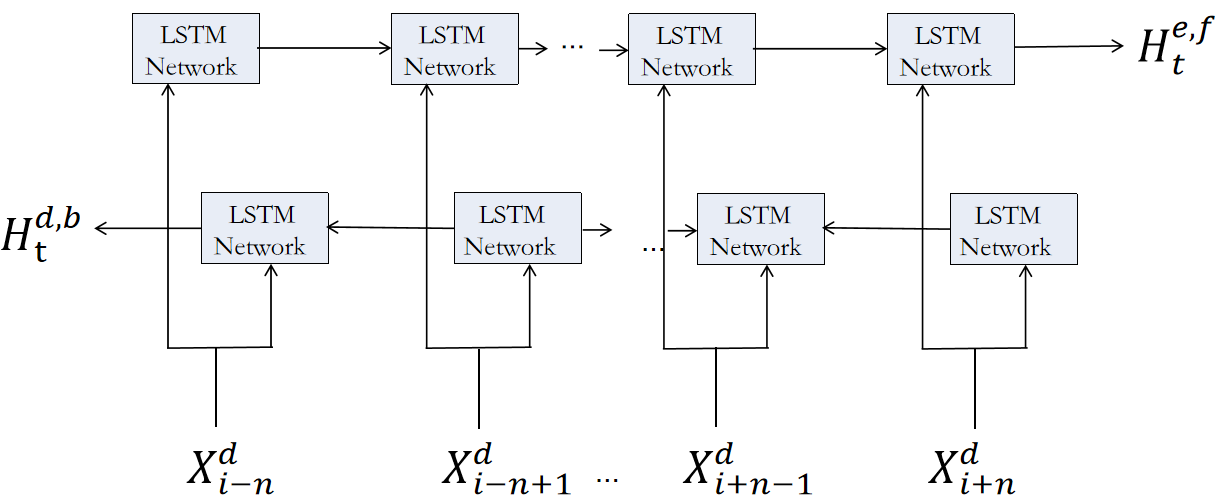}
    \caption{Bi-LSTM network structure}
    \label{fig:3}
\end{figure}

\section{Methodology}
    The magnetic core loss prediction model SEPI-TFPNet (Spectral Entropy–Physically Informed Temporal Feature Fusion Network) is shown in figure \ref{fig:1}. This method first uses an autoencoder to upscale the original input sequence in order to extract richer latent features. Then, it employs rotation embedding to encode positional information and introduces an attention mechanism to adaptively allocate weights to different time steps within the sequence. A feedforward neural network (FNN) is subsequently used to reconstruct the attention-weighted representations. The reconstructed results are standardized and normalized before being fed into a convolutional neural network (CNN), a bidirectional long short-term memory network (Bi-LSTM), and an empirical prior model (using the original data) to extract multi-scale, multi-modal features. Finally, the features from each modality are jointly encoded in an adaptive feature fusion layer and undergo final loss prediction via a multi-layer perceptron (MLP). The preliminary prediction results from the empirical prior model can also influence the final prediction through the loss function, thereby achieving dual constraints.

\subsection{Deep Learning Section}
This paper introduces autoencoders to fully explore the potential features of magnetic core data, thereby alleviating the constraints imposed by the limited number of original samples on CNN training. In the feature mapping and fusion stage, this paper uses a two-layer fully connected network (FNN) to perform nonlinear transformations on the features generated by the autoencoder.

Considering that magnetic flux sequences not only exhibit forward dependencies but also include backward influences of future values on the current state, this paper designs a feature extraction module based on a bidirectional LSTM (Bi-LSTM). By stacking forward and backward LSTMs, it achieves joint modeling of the global context of the sequence. In the output stage, by concatenating the bidirectional hidden states, it fully integrates contextual information from both directions, significantly enhancing the ability to capture periodic features.Figure \ref{fig:2} shows a single LSTM unit, and Figure \ref{fig:3} shows the structure of a bidirectional LSTM.

In the feature extraction stage, this paper designs a lightweight convolutional neural network (CNN) module to extract deep temporal information from feature maps generated by autoencoders and multi-head attention mechanisms. Unlike traditional CNNs combined with fully connected layers, this CNN abandons the fully connected structure with a large number of parameters and consists only of three layers of one-dimensional convolutions, two layers of max pooling, and one layer of flattening, in the following order: [Conv → MaxPool → Conv → MaxPool → Conv → Flatten].

The position encoding layer uses sine and cosine functions to encode the odd and even dimensions of the B(t) sequence, respectively, so that the data vector $\mathbf{x}_i$  at each time step is added to its corresponding position information $\mathbf{p}_i$, forming an embedding vector $\mathbf{x}'_i$ that includes position information. This preserves global position information in the 48-dimensional representation expanded by the autoencoder. Based on this representation, the model introduces a multi-head self-attention mechanism, dividing the B(t) sequence of 1024 sampling points into several subspaces and computing the scaled dot product attention in parallel.

To establish an accurate mapping between the feature vector and core loss prediction, this method uses a three-layer MLP to build the final core loss regression model.

\subsection{Feature fusion mechanism optimization}
In magnetic core loss modeling, traditional feature fusion methods often rely on simple concatenation or fixed-weight averaging, making it challenging to effectively incorporate the dynamic interdependencies of multi-source information across different operating conditions, thereby leading to performance bottlenecks. To address this limitation, this paper proposes an adaptive feature fusion (AFF) module within the SEPI-TFPNet framework, inspired by SENet \cite{30}.
The AFF module aims to autonomously learn the importance of each sub-feature stream based on the intrinsic feature distribution of the input data. Specifically, the time-series deep features  $F_t \in \mathbb{R}^{d_t}$ and preprocessed scalar features $F_s\in\mathbb{R}^{d_s}$are treated as two information streams, and fusion weights are generated through a gated attention mechanism:
\begin{equation}\alpha_t=\mathrm{softmax}(\mathbf{W}_t\cdot\phi(\mathbf{F}_t)+b_t),\end{equation}
\begin{equation}\alpha_s=\mathrm{softmax}(\mathbf{W}_s\cdot\phi(\mathbf{F}_s)+b_s),\end{equation}

Among them,$\phi(\cdot)$ denotes a one-dimensional global average pooling operation, \quad $W_t \in \mathbb{R}^{1 \times d_t}$, \quad $W_s \in \mathbb{R}^{1 \times d_s}$ and the bias terms bt, bs are learnable parameters. The final fusion feature representation is:

\begin{equation}\mathbf{F}_{\mathrm{AFF}}=\alpha_t\odot\mathbf{F}_t\oplus\alpha_s\odot\mathbf{F}_s,\end{equation}

Among them, “$\odot$” and “$\oplus$” represent channel-weighted and vector concatenation operations, respectively.

\subsection{Physical model based on spectral entropy discrimination}

This section proposes a hard switching strategy based on spectral entropy to improve the adaptability of the physical prior module.

For a given time-domain flux density waveform B(t), it is first discretized and its discrete Fourier transform is calculated to obtain the frequency-domain complex spectrum \(S(f_k)\):
\begin{equation}S[k]=F\{B[n]\}\end{equation}

Then, the power spectrum is calculated using the square amplitude:

\begin{equation}P(f_k)=\left|S(f_k)\right|^2\end{equation}

Then, normalize the power spectrum and treat it as a probability distribution:

\begin{equation}p_k=\frac{P(f_k)}{\sum_{i=1}^NP(f_i)},\quad\sum_{k=1}^Np_k=1.\end{equation}

Here, N is the number of frequency sampling points. The spectral entropy H is defined as the Shannon entropy of the normalized power spectrum:
\begin{equation}H=-\sum_{k=1}^Np_k\ln p_k,\end{equation}

The logarithm base can be the natural logarithm or other logarithm bases. Physically, spectral entropy characterizes the degree of disorder in the frequency spectrum of a magnetic flux waveform, which can be used to accurately distinguish between sinusoidal and non-sinusoidal waveforms.

Subsequently, the calculated spectral entropy H is compared with the threshold Hth to achieve hard switching between different loss models. Typically, the threshold \quad \(H_{\mathrm{th}}\) is set to distinguish waveform complexity: when \(H \le H_{\mathrm{th}}\), the input waveform spectrum is close to sinusoidal, and the traditional Steinmetz empirical formula is used for loss prediction; when  \quad \(H > H_{\mathrm{th}}\), the waveform spectrum is rich in higher-order harmonics, and the system switches to the more general iGSE model. The hard switching logic can be formalized as:
\begin{equation}P_{\mathrm{loss}}=
\begin{cases}
P_{\mathrm{Steinmetz}}(f,B_m), & H\leq H_{\mathrm{th}}, \\
P_{\mathrm{iGSE}}(B(t)), & H>H_{\mathrm{th}}, 
\end{cases}\end{equation}

The above strategy enables the model to automatically select a more suitable physical model based on spectral characteristics under different excitation waveforms, thereby improving the adaptability and accuracy of predictions and reducing errors. The $P_{\mathrm{Steinmetz}}$ formula is as follows:

\begin{equation}P_{v}=kf^{a}B_{m}^{b},\end{equation}

This formula is applicable to nearly sinusoidal excitation conditions, but its prediction accuracy is limited for non-sinusoidal waveforms. The improved generalized Steinmetz equation (iGSE) extends its applicability to arbitrary waveforms and is expressed as follows:

\begin{equation}P=\frac{1}{T}\int_{0}^{T}k_{i}\left|\frac{dB}{dt}\right|^{a}(\Delta B)^{b-a}dt,\end{equation}

The iGSE model can calculate losses under any waveform without requiring additional parameters, enabling more accurate reflection of loss characteristics under non-sinusoidal excitation conditions.

When switching models, the threshold \quad \(H_{\mathrm{th}}\) must be defined. Through experimental verification, the threshold \quad \(H_{\mathrm{th}}\) was determined to be 0.01.

\subsection{Custom loss function}
This paper designs a custom loss function composed of a double mean absolute percentage error (MAPE) term. Specifically, let \(P_{\mathrm{pred}}(x)\) be the predicted loss of the model under input condition x, \(P_{\mathrm{act}}(x)\)  be the corresponding measured loss value, and \(P_{\mathrm{emp}}(x)\) be the reference loss value calculated based on the classic physical empirical formula. Then, the loss function is defined as:

\begin{equation}
\resizebox{0.9\hsize}{!}{%
$\displaystyle
\mathcal{L}_{\mathrm{custom}}
= \lambda_{1}\,\frac{1}{N}\sum_{i=1}^{N}
  \frac{\bigl|P_{\mathrm{pred}}(x_{i})-P_{\mathrm{act}}(x_{i})\bigr|}
       {\bigl|P_{\mathrm{act}}(x_{i})\bigr|}
\;+\;
\lambda_{2}\,\frac{1}{N}\sum_{i=1}^{N}
  \frac{\bigl|P_{\mathrm{pred}}(x_{i})-P_{\mathrm{emp}}(x_{i})\bigr|}
       {\bigl|P_{\mathrm{emp}}(x_{i})\bigr|}$
}
\end{equation}

The first term is used to supervise the network to accurately restore data features. The second term serves as a physical consistency regularization term.The specific values of $\lambda_1$  and $\lambda_2$ are determined through grid search.

\section{Results and Discussion}

To validate the effectiveness of the core loss modeling method based on SEPI-TFPNet proposed in this paper, a series of relevant experiments were designed and conducted in this study. The data used in this study were obtained from the public magnetic materials database: MagNet\cite{8}.

\subsection{Comparison with the 2023 Magnet Challenge Model}

\begin{table}[!t]
  \centering
  \caption{Comparison of the model proposed in this paper with the results of the 2023 Magnet Challenge model}
  \label{tab:1}
  \begin{tabular}{lcc}
    \toprule
    \textbf{Team Name} & \textbf{Abs.95\%(\%)} & \textbf{Explainability} \\
    \midrule
    ASU & 5.6 & Low \\
    Fuzhou & 2.2 & Low \\
    HDU & 3.7 & Low \\
    KU-Leuven & 58 & Low \\
    NJUPT & 6.9 & Low \\
    NTU & 88.7 & Low \\
    NTUT & 7.4 & Low \\
    Tsinghua & 6.4 & Low \\
    UTK & 4.3 & Low \\
    XJTU & 3.8 & Low \\
    CU-Boulder & 7.8 & Middle \\
    IISc & 2.8 & Middle \\
    Paderborn & 2.2 & Middle \\
    PoliTO & 33.4 & Middle \\
    SAL & 138.7 & Middle \\
    SEU.WX & 12.9 & Middle \\
    Sydney & 3.7 & Middle \\
    Tribhuvan & 8 & Middle \\
    ZJUI & 6.1 & Middle \\
    Mondragon & 7.9 & High \\
    SEU-MC & 6.9 & High \\
    New Paderborn & 3.2 & Middle \\
    \textbf{Proposed method} & \textbf{2.04} & Middle \\
    \bottomrule
  \end{tabular}
\end{table}

To validate the superiority of the model, this paper compares it with the 21 competing models from the 2023 Magnet Challenge\cite{32} and the optimized version of New Paderborn released in 2025 \cite{31}. The model training uses T37 material data, and the comparison metrics are 95\% error and interpretability, which is divided into three levels: “High” for models based solely on physical formulas or empirical equations, ‘Middle’ for models that integrate data-driven modules with physical/empirical models, and “Low” for purely data-driven models. The New Paderborn is the optimized version released in 2025, which ranked first in performance in the competition. Table \ref{tab:1} presents the experimental results.

The results indicate that the method proposed in this paper outperforms all other models, demonstrating superior predictive accuracy; In terms of interpretability, the embedded empirical models provide the network with a transparent physical prior channel, combining high accuracy with strong interpretability. While it may slightly lag behind pure physical models in interpretability, it achieves the optimal trade-off between global error and interpretability, validating the practical value and potential for widespread application of the hybrid data-driven–physical prior paradigm in the field of core loss prediction.

\subsection{Comparison with advanced models}

\begin{table}[!t]
  \centering
  \caption{Comparison of the results of the model proposed in this paper with those of advanced models}
  \label{tab:2}
  \begin{tabular}{@{}lcc@{}}
    \toprule
    \textbf{Team Name}      & \textbf{Abs.\,95\% (\%)} & \textbf{Explainability} \\
    \midrule
    PI-MFF-CN               &  5.58                    & Middle                  \\
    New EMPINN              & 15.55                    & Middle                  \\
    New Paderborn           &  3.34                    & Middle                  \\
    
    \textbf{Proposed method}& \textbf{2.13}            & Middle                  \\
    \bottomrule
  \end{tabular}
\end{table}

Given the temporal lag of the 2023 model, this paper compares it with the 2024 and 2025 models to further validate its superiority. The model training uses N30 material data, and the comparison metrics are 95\% error and interpretability. PI-MFF-CN is the model released in 2024, New Paderborn is the model released in 2025, and New EMPINN is the empirical model released in 2025. Table \ref{tab:2} presents the experimental results.The experimental results indicate that the proposed model achieves a 36.3\% improvement in 95\% relative error compared to the New Paderborn model and an 86.4\% improvement compared to the New EMPINN model. The proposed model demonstrates significantly higher prediction accuracy than recently proposed methods. All four models employ a physics–data coupling hybrid framework and exhibit consistent performance in terms of interpretability.

\section{Conclusion}

This paper proposes a hybrid model, SEPI-TFPNet, for core loss modeling. The model consists of two submodels. Submodel A is a physical model that switches between empirical models based on spectral entropy to enhance generalization capabilities. Submodel B is a complex, sophisticated data-driven model that addresses the extraction of deep-level features and multi-modal feature fusion through a series of processes. A series of experiments on the MagNet dataset demonstrate the model's superiority.SEPI-TFPNet achieves robust and high-precision core loss prediction, marking an important step toward developing a general-purpose core loss predictor.


\bibliographystyle{IEEEtran}   
\bibliography{reference}           

\end{document}